\documentclass{article}

\usepackage{microtype}
\usepackage{graphicx}
\usepackage{booktabs}
\usepackage{hyperref}

\usepackage[preprint]{icml2026}

\usepackage{placeins}
\usepackage[T1]{fontenc}
\usepackage{amsmath,amssymb}
\usepackage{adjustbox}
\usepackage{xcolor}

\newcommand{\placeholder}[1]{%
  \noindent\fbox{\parbox{0.95\linewidth}{\centering\small\color{gray}%
  \textbf{PLACEHOLDER}\\#1}}}
\newcommand{\resulttable}[2]{%
  \IfFileExists{tables/#1.tex}{%
    \adjustbox{max width=\linewidth,max totalheight=0.85\textheight}{%
      \input{tables/#1.tex}}}{\placeholder{#2}}}

\providecommand{\figdir}{figs/}
\newcommand{\resultfig}[3]{%
  \IfFileExists{\figdir#1.pdf}{%
    \includegraphics[width=#2]{\figdir#1.pdf}}{\placeholder{#3}}}

\newcommand{\plr}{PLR$^\perp$}

\icmltitlerunning{Benchmarking UED on Quadrupeds}

\begin{document}

\twocolumn[
  \icmltitle{Is It the Curriculum or Just the Easy Start? \texorpdfstring{\\}{} Benchmarking Unsupervised Environment Design on Quadrupeds}

  \begin{icmlauthorlist}
    \icmlauthor{Jakub Gilewicz}{pwr,mk,solvro}\quad
    \icmlauthor{Dominik Galus}{pwr,mk,solvro}\quad
    \icmlauthor{Piotr Wojdas}{pwr,mk,solvro}
  \end{icmlauthorlist}
  \begin{center}
    \textsuperscript{1}Wroc{\l}aw University of Science and Technology\\
    Wybrze{\.z}e Wyspia{\'n}skiego 27, 50-370 Wroc{\l}aw, Poland\\
    \textsuperscript{2}Machinekind, Wroc{\l}aw, Poland\\
    \textsuperscript{3}Ko{\l}o Naukowe Solvro, Wroc{\l}aw University of
    Science and Technology, Wroc{\l}aw, Poland\\
    \texttt{\{275409,279690,279903\}@student.pwr.edu.pl}
  \end{center}

  \icmlkeywords{Unsupervised Environment Design, Curriculum Learning,
    Reinforcement Learning, Legged Locomotion, Benchmark}

  \vskip 0.15in
  \begin{center}
    {\includegraphics[width=\textwidth]{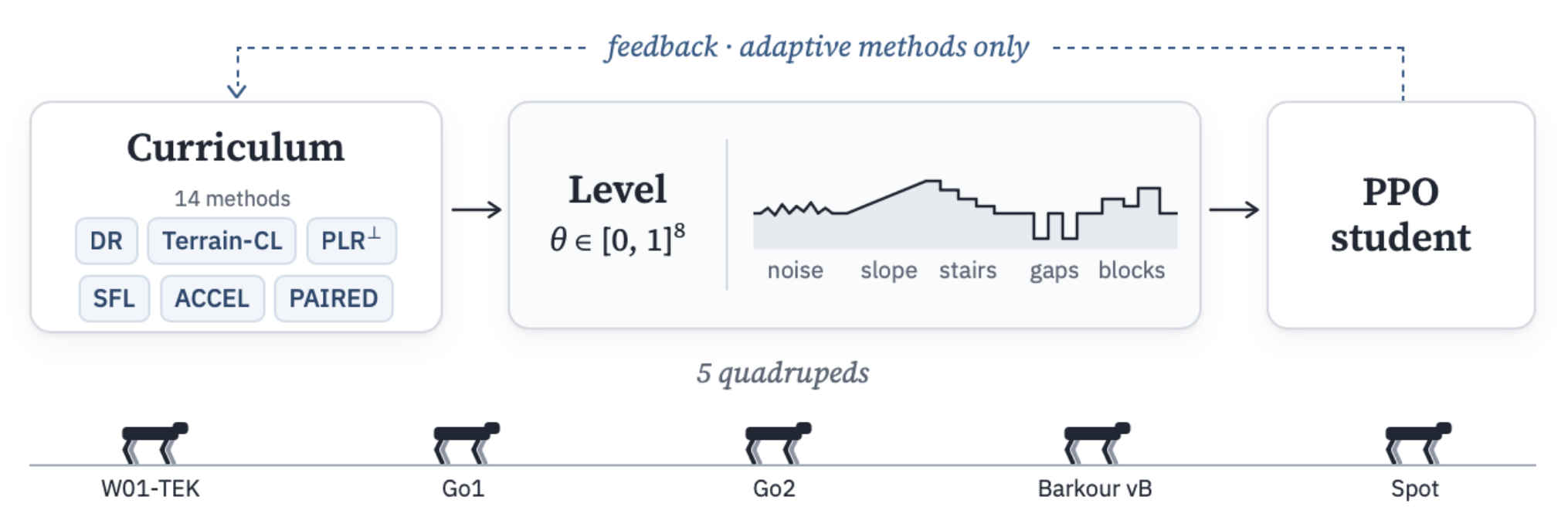}}
    \captionsetup{hypcap=false}
    \captionof{figure}{The benchmark. A curriculum proposes levels
    $\theta\in[0,1]^8$ of eight terrain parameters, and the same PPO
    student trains on them for each of five quadrupeds. Adaptive curricula
    choose the next levels from the student's rollouts (dashed).}
    \label{fig:teaser}
  \end{center}

  \vskip 0.2in
]

\makeatletter
{\let\thefootnote\relax\footnotetext{\hspace*{-\footnotesep}\Notice@String}}
\global\icml@noticeprintedtrue
\makeatother

\begin{abstract}
Unsupervised environment design (UED) methods such as PLR, ACCEL, SFL and
PAIRED are compared mostly on 2D mazes, grid worlds and BipedalWalker, while
terrain curricula for legged robots are hand-designed and evaluated against
a few baselines. We present a benchmark that runs fourteen UED,
domain-randomisation and robotics curricula under one protocol on five 3D
quadrupeds in MuJoCo MJX: the closed-chain W01-TEK and the open-chain
Unitree Go1 and Go2, Google Barkour vB and Boston Dynamics Spot. Levels are
vectors of eight interpretable terrain parameters, and all curricula drive the same PPO student
at matched environment-step budgets. The
evaluation reports success on random and held-out structured terrains,
interquartile means with stratified bootstrap intervals, capability
envelopes and a level-by-level domination comparison. We compare methods
within each robot and report, per level set, the top group of methods not
significantly below the best. No method is in the top group on every
robot and level set. Methods with easy or low-difficulty training levels,
curated or not, fill the top groups on in-distribution random levels of
Go1 and Barkour, and of W01-TEK apart from PAIRED+BiBC. The success-driven terrain curriculum from legged-robot practice
is in a top group most often, domain randomisation on uniform levels in
none, and PAIRED in none outside Spot, where the fourteen methods lie
within 3.8 points on random levels of difficulty 0.25. On three robots the leader on in-distribution random levels is
not the leader on held-out terrain. We collect the results into a per-robot
guide to choosing a curriculum.
\end{abstract}

\section{Introduction}
\label{sec:intro}

Learned locomotion controllers are trained on distributions of terrain,
and that distribution shapes what the controller can do. Practical
pipelines use hand-designed curricula that promote a robot along rows of
increasing difficulty~\citep{rudin2022learning} or adaptive samplers over a
parameterised terrain family~\citep{lee2020learning}. Recent work automates
this with learning-progress~\citep{li2026lpacrl},
learned-difficulty~\citep{liu2026tacl} and
safety-critic~\citep{ovinnikov2026psc} curricula, each with its own robot,
baselines and protocol.
Unsupervised environment design (UED) studies the same problem in general
form: a teacher proposes environment parameters (\emph{levels}) so that a
student generalises. PAIRED~\citep{dennis2020paired} learns a level
generator from a regret signal. PLR and its robust variant
\plr~\citep{jiang2021plr,jiang2021replay} curate random levels by a regret
score, ACCEL~\citep{parkerholder2022accel} mutates curated levels, and
SFL~\citep{rutherford2024noregrets} scores levels by learnability. These
methods are compared mostly on mazes, grid worlds, car racing and the 2D
BipedalWalker, with fast reference implementations in
minimax~\citep{jiang2023minimax} and JaxUED~\citep{coward2024jaxued}. To
our knowledge no benchmark runs the UED family under one protocol on 3D
legged locomotion, where the student is a contact-rich, high-dimensional
controller and a level is a terrain.

We build one and ask which UED findings carry over. Contributions:
\begin{itemize}
  \item \textbf{A five-robot benchmark} in MuJoCo
    MJX~\citep{todorov2012mujoco}: the closed-kinematic-chain quadruped
    W01-TEK and four open-chain robots from MuJoCo
    Playground~\citep{zakka2025playground} and
    Menagerie~\citep{menagerie2022github} (Go1, Go2, Barkour
    vB~\citep{caluwaerts2023barkour}, Spot), on a shared terrain level space.
  \item \textbf{Fourteen curricula on one student}: DR and two easy-start
    DR controls, the game-inspired terrain curriculum of
    \citet{rudin2022learning}, \plr, SFL, ACCEL with two mutation/score
    variants and a learning-progress score~\citep{li2026lpacrl}, and
    PAIRED with behaviour-cloning and evolutionary variants, all driving
    one brax PPO~\citep{schulman2017ppo,freeman2021brax} student at matched
    environment steps. Five seeds each, 350 runs.
  \item \textbf{An evaluation protocol}: success on random levels of
    graded difficulty and on held-out structured terrain, cross-fitted
    CVaR, aggregate interquartile means and probabilities of
    improvement~\citep{agarwal2021precipice}, capability envelopes and a
    level-by-level domination comparison with a binomial noise reference
    (\S\ref{sec:protocol}).
  \item \textbf{Findings and guide} (\S\ref{sec:results},
    \S\ref{sec:guide}), for one PPO student at matched environment
    steps, as answers to six questions asked per robot: which methods solve
    in-distribution random levels, which transfer to held-out terrain, which
    handle hard levels, which terrain features limit them, what they cost
    and how repeatable they are. No method is in the top group everywhere.
    Terrain-CL is in 13 of the 20 robot and level-set top groups, ACCEL in
    11 and DR on uniform levels in none. On Go1, Go2 and Barkour the leader
    on random levels is not the leader on held-out terrain. On Spot the
    fourteen methods lie within 3.8 points on $d=0.25$.
\end{itemize}
A method interacts with the benchmark only through the levels it proposes,
so new curricula plug in without changing the level space or evaluation.
The appendix holds the robot, level-space, method and protocol
details and all additional tables.
Code, level space and evaluation suite are available at
\url{https://github.com/Solvro/ml-wojtek-ued} under the MIT licence. The
W01-TEK model is available at \url{https://github.com/machinekind/w01-tek}
under the Apache 2.0 licence.

\section{Related Work}
\label{sec:related}

\paragraph{UED.} PAIRED~\citep{dennis2020paired} trains a generator
against the return gap of two students. \citet{mediratta2023stabilizing}
stabilise it with behaviour-cloning distillation between them.
\plr~\citep{jiang2021replay} trains only on replayed high-regret levels,
ACCEL~\citep{parkerholder2022accel} adds level edits and reports gains on
BipedalWalker, and SFL~\citep{rutherford2024noregrets} shows that common
regret estimates track success rate and replaces them with learnability.
We keep these algorithms close to their published form and change the
domain.

\paragraph{Locomotion curricula.} Terrain curricula for quadrupeds are
usually proposed one at a time and compared with a few
baselines~\citep{lee2020learning,rudin2022learning,li2026lpacrl,liu2026tacl,ovinnikov2026psc}.
The game-inspired curriculum of \citet{rudin2022learning}, which moves
each robot up or down a grid of terrain difficulties by the distance it
walked, is the de facto standard in legged-robot training pipelines.
LP-ACRL~\citep{li2026lpacrl} samples terrain tasks by learning progress,
the change in episodic return between consecutive training stages. We
include both, adapted to our level space (\S\ref{sec:benchmark}).
LLM-written courses~\citep{liang2024eurekaverse} and authored terrain
datasets~\citep{zhang2022terrain} also come with their own baselines. None runs the
UED family side by side under a shared level space and evaluation, and
none compares robots of different morphology and size on one
level space.

\section{Benchmark}
\label{sec:benchmark}

\paragraph{Robots and task.}
W01-TEK is a quadruped built at Wroc{\l}aw University of Science and
Technology, with 12 actuated
joints. Its legs close through a passive four-bar linkage (one
\texttt{connect} equality constraint per leg). The four open-chain robots
use MuJoCo Playground's joystick reward and observations on the default models of their manufacturers or
creators, with feet-only collisions and raised PD gains on Go2
(App.~\ref{app:robots}). Go2, which Playground lacks, runs Go1's
environment on the Menagerie model. Every robot tracks a velocity command
$(v_x, v_y, \omega_z)$ drawn from the forward half of its native command
box at a 0.02\,s control step. The actor is blind. The critic also sees a
$5\times5$ terrain height scan, so value-based level scores can tell
terrains apart. A robot falls when its trunk drops below 0.6 of its
standing height above the terrain beneath it or tilts past a gravity
threshold, the same rule for all five (Barkour also keeps Playground's
joint-limit check). Table~\ref{tab:robots} lists the
per-robot values. App.~\ref{app:robots} gives the full specification.

\paragraph{Success.}
Lengths are given in reference units and multiplied by each robot's size
factor $s$ (App.~\ref{app:robots}). An episode
\emph{succeeds} if the robot never falls and ends at least $s$ metres from
spawn. Relative to its size, Spot's top commanded speed is lower than the
other robots', so its episodes are longer (780 steps), which keeps the
success distance a similar share of the distance at top speed for all
robots
(App.~\ref{app:threshold}).

\begin{table*}[t]
  \centering\small
  \caption{Robots. Success = no fall and
  travel of at least $s$ metres within the episode.}
  \label{tab:robots}
  \begin{tabular}{llcc}
    \toprule
    Robot & Legs & Episode & Physics \\
    \midrule
    W01-TEK & closed chain (four-bar) & 300 steps & MJX \\
    Go1 & open chain &  300 steps & MJX \\
    Go2 & open chain & 300 steps & MJX \\
    Barkour vB & open chain &  300 steps & MJX \\
    Spot & open chain & 780 steps & MJX \\
    \bottomrule
  \end{tabular}
\end{table*}

\paragraph{Level space.}
A level is $\theta\in[0,1]^8$ plus a noise seed. The eight coordinates map
linearly to a radial slope ($-20^\circ$ to $+18^\circ$), radial stairs
($-8$ to $+5$\,cm high, 15-60\,cm wide), smooth noise (0-10\,cm
amplitude, 0.2-1\,m wavelength), gap trenches (0-6\,cm wide) and
piecewise-constant blocks (0-6\,cm high, 15-60\,cm cells), in reference
units (Table~\ref{tab:levelspace}). Features start at the edge of a flat spawn disc and are radial or
seeded fields, so a level does not depend on heading. Each parameter has a
\emph{flat} value: zero for slopes, heights and gaps, mid-range for widths,
cell sizes and wavelength, which have no effect at zero height. The level at \emph{difficulty} $d$ is
$\theta_d = \theta_{\text{flat}} + d\,(\theta - \theta_{\text{flat}})$. The
ranges are set on W01-TEK so that uniform levels are neither trivial
nor mostly unsolvable.
Levels are rendered into one shared heightfield: each of an iteration's
256 levels occupies its own tile, so every lane of 32 environments trains
on its own level without recompiling the model (App.~\ref{app:render}).

\paragraph{Student and training loop.}
All methods train the same student: brax PPO networks and loss with
MuJoCo Playground's locomotion hyperparameters (MLPs
$512\text{-}256\text{-}128$, ELU, observation normalisation, 20-step
unrolls, Table~\ref{tab:hparams}). One iteration resets 32 environments onto each of 256
levels, with a fresh physics-randomisation draw~\citep{tobin2017domain}
where the robot defines one, and
steps them for one episode, with a PPO update after every unroll when the
iteration trains. Runs last 300M environment steps for W01-TEK and 1B for
the other robots, five seeds per method and robot.

\paragraph{Methods.}
Table~\ref{tab:methods} lists the fourteen curricula. They differ in where
new levels come from (uniform, an \emph{easy} generator that shrinks a
uniform level to difficulty $d=0.1$, a per-level random difficulty, a
difficulty row, or a learned generator), whether a buffer of the 4000
highest-scoring levels is replayed by rank and staleness, whether replayed
levels are edited, and how levels are scored: positive value loss (mean
positive GAE advantage~\citep{jiang2021plr}), learnability $p(1-p)$ with
$p$ the lane's success fraction~\citep{rutherford2024noregrets}, learning
progress, or PAIRED regret (best antagonist return minus mean protagonist
return). DR-easy and DR-mixed are controls: they isolate ``train on easy
levels'' from curation.
\emph{Terrain-CL} adapts \citet{rudin2022learning} to the tiled arena:
each of the 256 lanes keeps one uniform level (its column) for the whole
run and a row $r\in\{0,\dots,10\}$, starting from $r\sim\mathcal{U}\{0..5\}$,
and trains on the column shrunk to $d=r/10$. After every iteration a lane
moves up a row when at least half of its environments succeed, down when
its mean travel is below half the commanded displacement, and to a random
row when it passes the top.
\emph{ACCEL-LP} scores ACCEL's levels by learning
progress~\citep{li2026lpacrl}: a visit's score is the lane's mean
first-episode return minus the return of the level's previous visit,
signed, and 0 on a level's first visit.
All methods get the same number of environment steps. Non-training
scoring iterations count toward it, and PAIRED's antagonist does not, so
PAIRED simulates twice as much (App.~\ref{app:methods}).

\begin{table*}[t]
  \centering\small
  \caption{Curricula. Easy: uniform level shrunk to $d=0.1$.
  Mixed: per-level $d\sim\mathcal{U}(0,1)$. Rows: a fixed uniform level
  per lane at $d=r/10$, $r$ promoted or demoted each iteration. Edit: perturb 2 parameters by
  $\mathcal{U}(-0.1,0.1)$. Grow: also scale the distance from flat by
  $\mathcal{U}(1,1.3)$. PVL: positive value loss.}
  \label{tab:methods}
  \begin{tabular}{lllll}
    \toprule
    Method & New levels & Replay & Edit & Score \\
    \midrule
    DR~\citep{tobin2017domain} & uniform & - & - & - \\
    DR-easy & easy & - & - & - \\
    DR-mixed & mixed & - & - & - \\
    Terrain-CL~\citep{rudin2022learning} & rows & - & - & - \\
    \plr~\citep{jiang2021replay} & uniform & buffer & - & PVL \\
    SFL$^\dagger$~\citep{rutherford2024noregrets} & uniform & buffer & - & learnability \\
    ACCEL~\citep{parkerholder2022accel} & easy & buffer & edit & PVL \\
    ACCEL-grow & easy & buffer & edit + grow & PVL \\
    ACCEL-SFL & easy & buffer & edit + grow & learnability \\
    ACCEL-LP & easy & buffer & edit & learning progress \\
    PAIRED~\citep{dennis2020paired} & generator & - & - & PAIRED regret \\
    PAIRED+UniBC~\citep{mediratta2023stabilizing} & generator & - & - & PAIRED regret \\
    PAIRED+BiBC~\citep{mediratta2023stabilizing} & generator & - & - & PAIRED regret \\
    PAIRED+Evo & easy & buffer & edit & PAIRED regret \\
    \bottomrule
  \end{tabular}\\[2pt]
  {\footnotesize $^\dagger$\plr\ with the learnability score, not SFL's
  periodically rebuilt buffer (App.~\ref{app:methods}).}
\end{table*}

\section{Evaluation Protocol}
\label{sec:protocol}

Each final policy runs deterministically, without physics randomisation,
with commands drawn as in training except that the forward speed $v_x$
is high enough to reach the success distance. Every
level is rolled out in two independent passes of four episodes.

\paragraph{Level sets.}
(i)~\emph{Random}: 2048 uniform genomes at difficulty
$d\in\{0.25,0.5\}$, the same genomes at both, and a second, independent
draw of 2048 uniform genomes at full difficulty.
(ii)~\emph{Held-out suite}: seven structured terrain families no method
trains on (stairs up and down, slopes up and down, trenches, waves, random
blocks), each at four strengths with 64 instances.
(iii)~\emph{Flat}: no terrain, a check that terrain skills were not bought
with basic walking. (iv)~\emph{Capability envelopes}: one feature ramped
to the end of its range with the others flat. The envelope is the largest
magnitude passed with 50\% success (App.~\ref{app:protocol}).

\paragraph{Headline difficulty.} For the open-chain robots
full-difficulty levels are rarely solved: the best method solves 19\% of
them on Go1 and at most 5.0\% on Go2, Barkour and Spot, against 43\% on
W01-TEK. We therefore report $d=0.25$
as the headline set and full-difficulty results in App.~\ref{app:uniform}.

\paragraph{Cross-fitted CVaR.} Robustness is the mean success over the
worst 25\% of levels. Ranking and scoring levels on the same episodes
biases this low through selection noise (with 2000 levels of true success
0.5 and 4 episodes each, same-pass CVaR$_{25}$ reads 0.19), so levels are
ranked on one pass and scored on the other, in both directions.

\paragraph{Domination comparison.} Mean success hides whether one method
is better everywhere or the two win on different levels. For methods A and
B, each level's solve rate is pooled over all seeds and both passes
($5\times8=40$ episodes). We report the fraction of levels on which A's
rate is higher, lower or tied, the mean per-level difference with a 95\%
bootstrap interval, and the mean absolute difference against a binomial
noise reference: the mean $|A-B|$ that two identical policies would show
from episode sampling alone.

\paragraph{Statistics.} Tables report the mean $\pm$ standard deviation
over five seeds. Methods are compared within each robot with two-sided
Welch $t$-tests over the per-seed means. A robot's \emph{top group} on a
level set is the method with the best mean and every method not
significantly below it ($p\ge0.05$). Aggregates over robots (App.~\ref{app:results}) follow
\citet{agarwal2021precipice}: each robot's per-seed scores are min-max
normalised over the method means (0 = worst method, 1 = best), and we
report the interquartile mean (IQM) over the 25 robot-seed scores with a
95\% stratified bootstrap interval (robots as strata), and the
probability of improvement $P(X>Y)$, the Mann-Whitney probability that
a seed of $X$ beats a seed of $Y$ on the same robot, averaged over
robots. We call a difference significant at $p<0.05$, without
multiple-comparison correction, and give $p$-values for the pairwise
comparisons cited in the text.

\section{Results}
\label{sec:results}

\begin{table*}[t]
  \centering\small
  \caption{Success rate (\%) of the final deterministic policy, mean
  $\pm$ std over 5 seeds. Bold: the robot's top group on that level set,
  the best mean and every method not significantly below it (Welch
  $t$-test, $p\ge0.05$). $d{=}0.25$: 2048 random levels at difficulty
  0.25. OOD suite: held-out structured terrains. CVaR, $d=0.5$, flat
  ground and full-difficulty levels are in App.~\ref{app:results}.}
  \label{tab:main}
  \resulttable{main}{Main results table (\texttt{tabs/main.tex}).}
\end{table*}

\begin{figure*}[t]
  \centering
  \resultfig{curves}{\linewidth}{Learning curves
  (\texttt{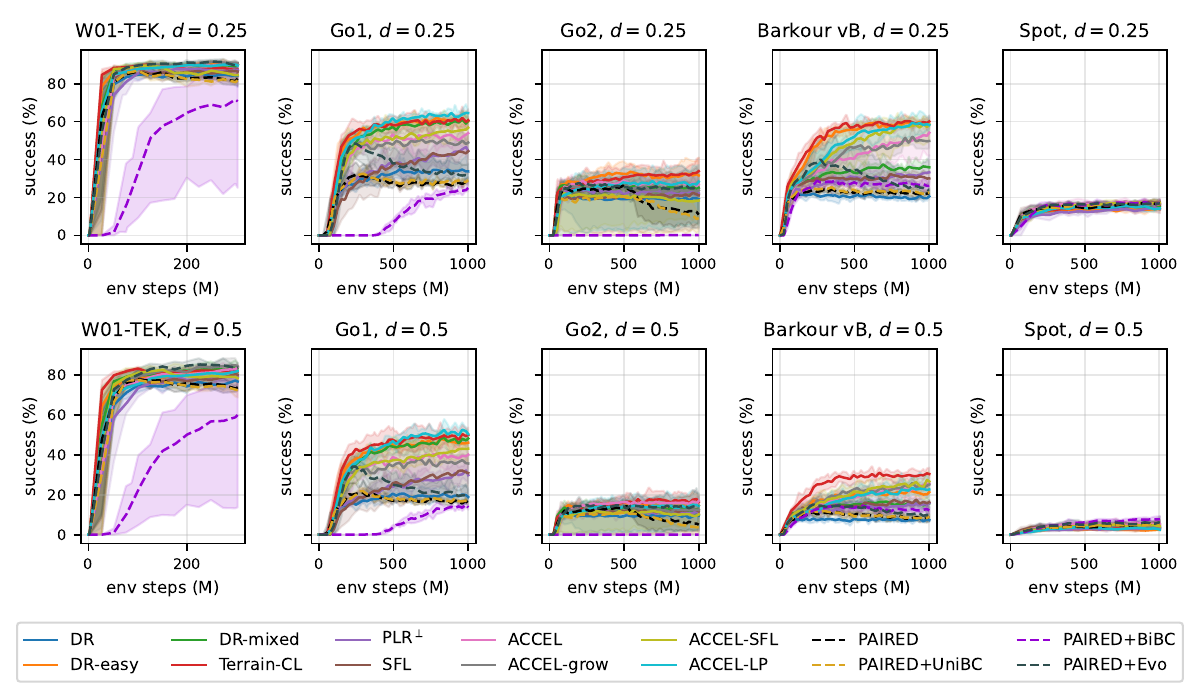}).}
  \caption{Success of the deterministic policy on 512 random levels at
  $d=0.25$ (top) and $d=0.5$ (bottom) during training. Mean over 5 seeds,
  shaded min-max. PAIRED and its variants dashed.}
  \label{fig:curves}
\end{figure*}

We answer six questions, each per robot. The same levels are not equally
hard for every robot (DR solves 84.5\% of $d=0.25$ levels on W01-TEK and
14.9-34.9\% on the others), so we compare methods within a robot and do
not pool rankings across robots. A method is in a robot's \emph{top
group} on a level set when its mean is the best or not significantly
below the best (Welch $t$-test, $p\ge0.05$, 5 seeds). A method outside
the top group is significantly worse than the best at uncorrected
$p<0.05$, while a method inside it is not shown to be equal. Membership
also depends on the seed spread: on W01-TEK's $d=0.25$ levels
PAIRED+BiBC ($70.3\pm26.4\%$) is in the top group and DR-easy
($90.2\pm0.7\%$) is not. Table~\ref{tab:guide} (\S\ref{sec:guide}) collects the top groups
of all level sets. Aggregates over robots are in App.~\ref{app:results}.

\paragraph{Q1: Which methods solve in-distribution random levels?} On
W01-TEK all methods except PAIRED+BiBC solve 80.5-91.2\% of $d=0.25$
levels, and the top group is ACCEL-grow (91.2\%), ACCEL, DR-mixed and
PAIRED+BiBC. On
Go1 it is ACCEL-LP (64.0\%), Terrain-CL, DR-easy and DR-mixed, with DR at
34.9\% and the PAIRED family at 24.6-33.3\%. On Go2 it is Terrain-CL
(34.1\%), DR-easy, ACCEL-LP, ACCEL and SFL. On Barkour it is Terrain-CL
(60.1\%), ACCEL-SFL, DR-easy, ACCEL-LP and ACCEL. On Spot the fourteen
methods lie within 14.1-17.9\%, a spread of 3.8 points, and the top
group is PAIRED+BiBC, PAIRED, ACCEL, ACCEL-SFL and PAIRED+Evo. On Go1
and Barkour every member of the top group uses easy or low-difficulty
training levels (the easy generator, mixed difficulty or Terrain-CL's low
rows), and on W01-TEK every member except PAIRED+BiBC does. The methods
that train on the full uniform range (DR, \plr, SFL) are outside it on
all three robots, and so are the learned-generator methods apart from
PAIRED+BiBC on W01-TEK. At
$d=0.5$ the top group is a subset of the $d=0.25$ group on W01-TEK, Go1,
Barkour and Spot. On Go2 it widens to eight methods, because seeds that never
learn to walk inflate the variance (Q6).

\paragraph{Q2: Which methods transfer to held-out terrain?} On the
held-out suite the top group of W01-TEK is ACCEL-grow (59.0\%),
Terrain-CL, DR-mixed, PAIRED+Evo and ACCEL. On Go1 it is Terrain-CL
(45.8\%), ACCEL and SFL, on Go2 ACCEL (33.4\%), SFL, ACCEL-LP and
ACCEL-SFL, on Barkour SFL (45.2\%) and Terrain-CL, and on Spot
PAIRED+BiBC (23.2\%) and PAIRED+Evo. On Go1, Go2 and Barkour the leader on
$d=0.25$ is not the leader on the suite. Methods that tie on random levels
can differ widely on held-out terrain: DR-easy and Terrain-CL solve the
same share of Go1's $d=0.25$ levels (61.6 vs.\ 62.3\%) but differ by 7.2
points on the suite (38.6 vs.\ 45.8\%), and ACCEL-LP and Terrain-CL tie
on Barkour's $d=0.25$ levels (59.0 vs.\ 60.1\%) and differ by 12.7 points
on its suite (30.1 vs.\ 42.8\%). SFL shows the reverse: it is outside the $d=0.25$ top group on
Go1 and Barkour and in the suite's top group on both.

\paragraph{Q3: Which methods handle hard levels?} Uniform levels at full
difficulty separate the methods on W01-TEK and Go1, where the best
method, Terrain-CL in both cases, solves 43.5\% and 19.1\%. On Go2,
Barkour and Spot no method exceeds 5.0\% (Table~\ref{tab:uniform}). Terrain-CL is in the uniform-level top group on
four robots: with DR-mixed, ACCEL-grow, PAIRED+BiBC and PAIRED+Evo on
W01-TEK, with
ACCEL-LP and ACCEL on Go1, with ACCEL on Go2 and alone on Barkour. DR,
which trains on exactly these levels, is in the top group on no robot. On
the hardest quarter of the $d=0.25$ levels (CVaR$_{25}$) the top group is
the mean's top group or a subset of it on every robot except Spot, so on
these robots the methods that solve more levels do not do so by giving up
their hardest ones (Table~\ref{tab:detail}).

\paragraph{Q4: Which terrain features limit each method?} Capability
envelopes (Table~\ref{tab:envelopes}, reference units) show where each
method stops. On W01-TEK all methods except PAIRED+BiBC reach the end of
the range on rough ground, downhill slopes and downward stairs, and stair-up
height separates them (5.0\,cm for DR-mixed, Terrain-CL, ACCEL,
ACCEL-grow, ACCEL-SFL and PAIRED+Evo, 4.0\,cm for DR). On Go1 stair-up
height (Terrain-CL 4.1, ACCEL 3.9, SFL 3.6\,cm) and gaps (at most
2.7\,cm, ACCEL-LP) bind. On Barkour SFL alone reaches the best block
height (4.0\,cm) and shares the stair-up top group with PAIRED+Evo. On Go2
and Spot no method passes any gap. Spot's uphill-slope envelope is at most $5.2^\circ$.

\paragraph{Q5: How fast do methods learn, and at what cost?} On W01-TEK
every method but PAIRED+BiBC plateaus within about 100M steps
(Figure~\ref{fig:curves}). Go2 and Spot plateau early at low success. On
Go1 and Barkour the curves are still rising at 1B steps, so their
rankings hold at this budget rather than at convergence. At matched
environment steps every method outside the PAIRED family takes about
the same wall clock on a given robot (0.5-0.9 GPU-hours per run,
Table~\ref{tab:compute}). Buffer methods spend a fifth to a quarter of
their steps on scoring iterations that do not update the student (786M student
steps of 1B for ACCEL on Go1). PAIRED and its variants simulate twice
as many steps, for the antagonist, and take 1.8-1.9 times the wall clock.

\paragraph{Q6: How repeatable are the results across seeds?} Runs that
never learn to walk concentrate on Go2, where
standard deviations reach 12.7 points (ACCEL-LP, $d=0.25$) and the top
groups widen. Outside Go2 the largest standard deviations on $d=0.25$ and
the suite, apart from PAIRED+BiBC on W01-TEK, are SFL's on Go1 (8.2 and
4.4), ACCEL's on Barkour's $d=0.25$ levels (7.2) and PAIRED+Evo's on
Barkour's suite (7.8). Terrain-CL stays at or below 3.1 points outside
Go2, and DR-mixed at or below 2.8 on every robot.

\section{Which Curriculum Should I Use?}
\label{sec:guide}

\begin{table*}[t]
  \centering\small
  \caption{Top-group membership per robot and level set ($\bullet$: the
  best mean or not significantly below it, Welch $t$-test, $p\ge0.05$, 5
  seeds). $.25$, $.5$: random levels at $d=0.25$, $0.5$. OOD: held-out
  suite. U: uniform levels at full difficulty. On Spot all methods lie within 3.8 points
  on $d=0.25$.}
  \label{tab:guide}
  \resulttable{guide}{Top-group table (\texttt{tabs/guide.tex}).}
\end{table*}

Table~\ref{tab:guide} marks, per robot and level set, the methods in the
top group. We summarise it per family of methods. Every statement holds
for one PPO student at this budget.

\paragraph{Randomisation (DR, DR-easy, DR-mixed).} DR on uniform levels
is in no top group. Changing only the distribution it samples from
changes this. DR-easy is in the $d=0.25$ top group on Go1, Go2 and Barkour
and in no held-out top group. DR-mixed is in every top group on W01-TEK
and in both random-level groups on Go1. Both cost the same as DR and need
no buffer or score.

\paragraph{Terrain curriculum (Terrain-CL).} Terrain-CL is in more top
groups than any other method (13 of 20): all four on Go1 and Barkour,
three on Go2, the suite and uniform levels on W01-TEK, and none on Spot.
It needs the task's success test and legged\_gym's travel threshold, and
no buffer or score. Its rows need no robot-specific
setting.

\paragraph{Replay of uniform levels (\plr, SFL).} \plr\ is in no top
group, but on Go1 and Barkour it improves on DR, which samples the same
levels (45.9 vs.\ 34.9\%, $p=0.002$, and 33.5 vs.\ 20.9\%, $p=0.003$, at
$d=0.25$). On W01-TEK, Go2 and Spot it does not. SFL, the same buffer
scored by learnability, is in the held-out top group on Go1, Go2 and
Barkour, leads it on Barkour, and has a wide seed spread on Go1 and Go2 (std up
to 12.2 points on $d=0.25$ and the suite).

\paragraph{ACCEL and its variants.} ACCEL is in the held-out top group on
W01-TEK, Go1 and Go2, leads it on Go2, and in the $d=0.25$ group on every
robot except Go1, eleven groups in total. ACCEL-grow, which also enlarges
edited levels, leads W01-TEK's $d=0.25$, $d=0.5$ and held-out sets and enters no other robot's group
except Go2's at $d=0.5$. ACCEL-LP leads Go1's random levels and is in
Go2's held-out group, but in no other held-out group. ACCEL-SFL is in
Barkour's random-level groups, Go2's held-out group and Spot's $d=0.25$
group, and is below ACCEL on W01-TEK (83.8 vs.\ 90.5\%, $p=0.002$).

\paragraph{PAIRED and its variants.} PAIRED is in no top group outside
Spot's $d=0.25$ set, and below \plr\ on W01-TEK, Go1, Go2 and Barkour (on
Go1 28.9 vs.\ 45.9\%, $p<0.001$). Distilling the antagonist into the
protagonist (UniBC) changes PAIRED's $d=0.25$ success by at most 1.3
points on any robot. Distilling both ways (BiBC) collapses on Go2 (0.2\%
success) and leads every Spot set. PAIRED+Evo,
which replaces the learned generator with ACCEL's buffer and edits under
the same regret, is in the held-out top group on W01-TEK and Spot. All
four cost about twice the wall clock of the other methods.

\paragraph{Short answers.} For in-distribution random levels, a method
with easy or low-difficulty training levels is in the $d=0.25$ top group on
every robot, whether it curates (ACCEL family, Terrain-CL) or not (DR-easy,
DR-mixed). For held-out terrain the choice depends on the robot: Terrain-CL or
ACCEL on Go1, ACCEL on Go2, SFL or Terrain-CL on Barkour, PAIRED+BiBC
or PAIRED+Evo on Spot, and ACCEL or ACCEL-grow, Terrain-CL, DR-mixed or
PAIRED+Evo on W01-TEK. Without a
budget to compare methods on a new robot, Terrain-CL is the method found
in a top group most often here, and ACCEL is second. Evaluating only on
in-distribution random levels would have picked a different leader on Go1,
Go2 and Barkour.

\section{Discussion and Limitations}
\label{sec:discussion}

\paragraph{Relation to 2D UED results.} On BipedalWalker ACCEL outperforms
\plr, PAIRED and DR~\citep{parkerholder2022accel}. Here ACCEL is in
eleven of twenty top groups, and Terrain-CL, a curriculum from
legged-robot practice, is in thirteen. JaxUED and SFL report DR as a
strong baseline in 2D domains~\citep{coward2024jaxued,
rutherford2024noregrets}. Here DR on uniform levels is in no top group,
while DR on easy or mixed-difficulty levels is in the random-level top
groups of four robots. PAIRED trails replay-based methods in
2D~\citep{parkerholder2022accel,jiang2021replay} and here, and the
behaviour-cloning fixes proposed for it~\citep{mediratta2023stabilizing}
do not change that outside Spot.

\paragraph{Benchmark statement.} The benchmark fixes one training budget per
robot, shared by all methods. Training can be extended, and longer runs
may improve the policies. On three robots the leader on in-distribution
random levels is not the leader on held-out terrain, so a new curriculum should be reported on
both.

\paragraph{Limitations.}
\begin{itemize}
  \item A robot's structure and build affect the share of levels it
    solves, whatever the method. We therefore compare methods within
    each robot, and the aggregates over robots (App.~\ref{app:results})
    average scores normalised per robot rather than raw success.
  \item Terrain-CL and ACCEL-LP are adaptations: Terrain-CL's columns are
    uniform levels of the level space rather than legged\_gym's terrain types,
    and promotion is decided per lane of 32 environments. ACCEL-LP ranks a
    replay buffer by learning progress measured over one visit, instead of
    LP-ACRL's softmax over a fixed task set.
\end{itemize}

\paragraph{Conclusion.} Across fourteen curricula and five quadrupeds no
method is in the top group on every robot and level set. Methods with
easy or low-difficulty training levels, curated or not, fill the
random-level top groups of Go1 and Barkour, and of W01-TEK apart from
PAIRED+BiBC. The success-driven
terrain curriculum is in a top group most often and ACCEL second, while
DR on uniform levels is in none and PAIRED in none outside Spot. The
leader on held-out terrain differs from the leader on in-distribution
random levels on three robots. The benchmark, level space and protocol
are built so that a new curriculum is added by changing only how levels
are proposed.

\section*{Acknowledgments and Disclosure of Funding}
Calculations have been carried out in Wroc{\l}aw Centre for Networking and
Supercomputing (\url{http://www.wcss.pl}), grant
No.~hpc-danbor2008-1756464546. W01-TEK was built in the Laboratory of
Mechatronics and Robotics, Wroc{\l}aw University of Science and
Technology, under the scientific supervision of Jaros{\l}aw Szrek. We
thank Machinekind for providing its robot model.

\section*{Impact Statement}
This paper presents work whose goal is to advance the field of Machine
Learning. There are many potential societal consequences of our work, none
of which we feel must be specifically highlighted here.
The authors hope that this benchmark and its per-robot guide will
support the development of Physical AI and of learning-based robotics,
by making curricula for legged robots easier to compare, reproduce and
choose.

\bibliography{refs}
\bibliographystyle{icml2026}

\newpage
\appendix
\onecolumn


\FloatBarrier
\section{Robots and Task}
\label{app:robots}

\FloatBarrier
\subsection{W01-TEK}
W01-TEK is a quadruped built in the Laboratory of Mechatronics and
Robotics, Wroc{\l}aw University of Science and Technology. Machinekind
provided the robot model we simulate
(\url{https://github.com/machinekind/w01-tek}). It has 12 actuated joints (three
per leg). Each leg is a
closed kinematic chain: the actuated joints drive the foot through a
passive four-bar linkage, closed in MuJoCo by one \texttt{connect}
equality constraint per leg. The closure adds constraint dynamics that
serial legs do not have (the linkage can reach a toggle configuration,
which the reward penalises). Training randomises terrain friction, base
and link masses, and actuator gain and damping scale per environment.
W01-TEK runs on the MuJoCo Warp backend of MJX. The four Playground robots
run on the JAX backend.

\FloatBarrier
\subsection{Open-chain robots}
Go1, Barkour vB and Spot use MuJoCo Playground's joystick
environments~\citep{zakka2025playground}: Playground's reward and
observations, feet-only collisions and, where Playground defines one, its
physics randomisation. Go2 has no Playground environment. It runs Go1's
environment on the Menagerie model~\citep{menagerie2022github}, rewritten
to feet-only collisions and the sensor names Go1's environment reads. Every
robot keeps its own reward. We change the scene (the tiled heightfield
arena), the commands (forward-walking slice of the native command box),
the termination (the shared fall rule below) and the critic input (a
height scan). Table~\ref{tab:robots-full} lists the per-robot values.

\begin{table}[!htbp]
  \centering\small
  \caption{Per-robot settings.}
  \label{tab:robots-full}
  \begin{tabular}{llllll}
    \toprule
    & W01-TEK & Go1 & Go2 & Barkour vB & Spot \\
    \midrule
    Model & Machinekind & Playground & Menagerie & Menagerie & Playground \\
    Episode (steps) & 300 & 300 & 300 & 300 & 780 \\
    Physics DR & own & Playground & Go1's & none & none \\
    \bottomrule
  \end{tabular}
\end{table}

\paragraph{Models.} Every robot keeps the default actuator and joint
settings of the model released by its manufacturer or creators (the
MJCF/URDF files of MuJoCo Menagerie and MuJoCo Playground, and
Machinekind's model of W01-TEK). Only Go2's PD gains are raised
(position gain doubled, velocity gain and joint damping scaled by
$\sqrt2$, keeping Menagerie's damping ratio).

\paragraph{Size.} $s$ is set per robot from its settled standing pose over
flat ground, relative to a fixed reference, and rounded to 0.1. It scales the
arena, every level length, the eval suite, the critic's height-scan grid
and the success distance. Slopes and the fall rule's ratio do not scale.

\paragraph{Fall rule.} An episode ends in a fall when the trunk is below
$0.6$ of the standing height above the terrain under it (read from the
heightfield) or the gravity vector's $z$-component in the trunk frame
exceeds $-0.4$. This replaces Playground's own terminations, which use
world heights (wrong on a pit or a raised tile). Barkour keeps
Playground's joint-limit check.

\paragraph{Critic.} The critic additionally receives a clean $5\times5$
terrain height scan ahead of the robot (a fixed grid scaled by $s$). The
actor is blind.

\FloatBarrier
\subsection{Success distance and episode length}
\label{app:threshold}
Success needs $s$ metres of travel within the episode. With 300-step
(6\,s) episodes, this is 28-30\% of the distance at top commanded speed
for W01-TEK, Go1, Go2 and Barkour. Relative to its size, Spot's top
commanded speed is lower than the other robots', so in 6\,s its success
distance would need 72\% of its top speed from rest. With 6\,s episodes
even a perfect flat walker would succeed on about half of the levels. We
lengthen Spot's episodes to 780 steps (15.6\,s), which restores the 28\%
ratio and
keeps the success distance a size-relative quantity. Evaluation draws
$v_x$ from $1.2\times$ the success distance over the episode duration up
to the maximum, so every evaluation command can reach the distance. The
distance is then 83\% of the travel at the slowest evaluation command for
all five robots. Spot redraws its command every 200 steps, Go1 and Go2
every 5\,s on average, and Barkour keeps one command per episode. With turning, perfect tracking of
evaluation commands reaches the distance on 62\% of Spot's levels, 70\%
of Go1's and Go2's and 95\% of Barkour's, which approximately caps
flat-ground success (Table~\ref{tab:detail}).

\FloatBarrier
\section{Level Space and Arena}
\label{app:levels}

\FloatBarrier
\subsection{Level space}
\begin{table}[!htbp]
  \centering\small
  \caption{Level space, reference units. Every length is multiplied by
  the robot's size $s$. A level is $\theta\in[0,1]^8$ plus a noise seed.}
  \label{tab:levelspace}
  \begin{tabular}{ll}
    \toprule
    Parameter & Range \\
    \midrule
    Noise amplitude & 0-0.10\,m \\
    Noise wavelength & 0.2-1.0\,m \\
    Radial slope & $-20^\circ$ to $+18^\circ$ \\
    Stair height & $-0.08$ to $+0.05$\,m \\
    Stair width & 0.15-0.60\,m \\
    Gap width & 0-0.06\,m \\
    Block height & 0-0.06\,m \\
    Block size & 0.15-0.60\,m \\
    \bottomrule
  \end{tabular}
\end{table}
Each parameter has a \emph{flat} value (zero slope, step, amplitude, gap
and block height, and mid-range stair width, noise wavelength and block
size, which have no effect at zero height). Because the slope and stair ranges are asymmetric, flat
is not the centre of the cube. Difficulty $d$ moves a genome toward flat,
$\theta_d=\theta_{\text{flat}}+d(\theta-\theta_{\text{flat}})$. The easy
generator of DR-easy and ACCEL is $d=0.1$.

\FloatBarrier
\subsection{Rendering and the tiled arena}
\label{app:render}
A level is rendered into a square patch of half-width 3.0\,m at 0.05\,m
resolution with heights clipped to $\pm0.3$\,m (all $\times s$). A flat
spawn disc of radius $r_0=0.55$\,m blends into the terrain over $w=0.2$\,m.
Let $r$ be the distance from the patch centre, $\delta=\max(r-r_0-w,0)$ and
$\tau(r)=\mathrm{clip}((r-r_0)/w,0,1)$. Then
\begin{equation*}
  h(x,y) = \tau(r)\Big[\tan(\phi)\,\delta
    + h_s\Big\lceil\frac{\delta}{w_s}\Big\rceil
    + A\,n_\lambda(x,y) + h_b\,b_c(x,y)
    - D\,\mathbf{1}\big[\delta>0,\ (\delta \bmod P) < g\big]\Big],
\end{equation*}
where $\phi$ is the slope, $h_s,w_s$ stair height and width, $A,\lambda$
noise amplitude and wavelength, $n_\lambda$ a seeded bilinear noise field
in $[-1,1]$ with control points every $\lambda$, $b_c$ a seeded
piecewise-constant field in $[-1,1]$ with cells of size $c$ and height
$h_b$, and $g$ the gap width with depth $D=0.15$\,m and period $P=0.6$\,m.

\begin{figure}[!htbp]
  \centering
  \resultfig{terrains}{\linewidth}{Rendered example levels
  (\texttt{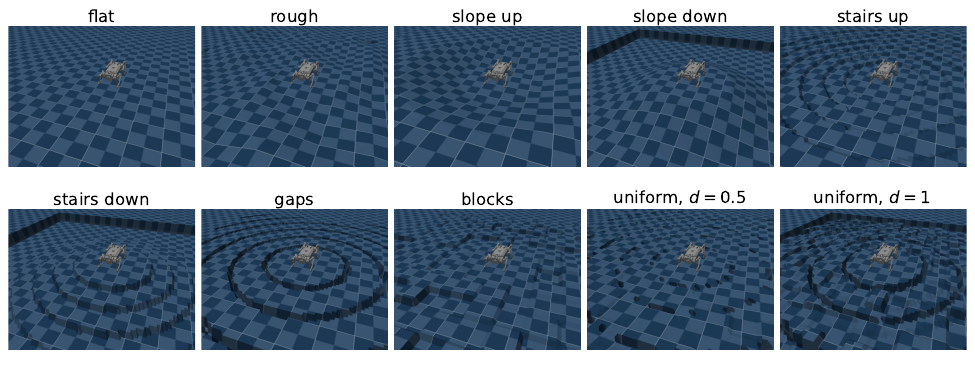}).}
  \caption{W01-TEK on example levels. The first panel is flat. Each of the
  next seven sets one feature to the end of its range with every other
  feature flat. The
  last two are uniform samples at $d=0.5$ and $d=1$. Features start at the
  edge of the flat spawn disc.}
  \label{fig:terrains}
\end{figure}

\begin{figure}[!htbp]
  \centering
  \resultfig{robots}{\linewidth}{The five robots on one level
  (\texttt{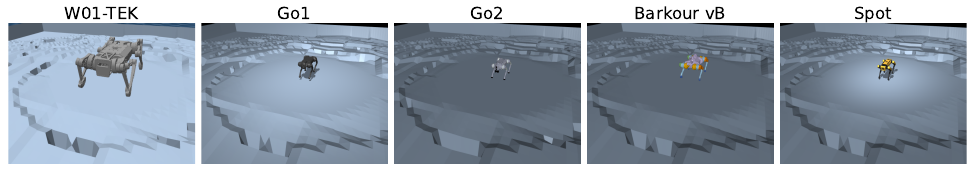}).}
  \caption{One level genome ($d=0.6$) rendered for each robot in its own
  arena. Terrain lengths and camera distance scale with the robot's size
  $s$, so robots of different proportions fill their panels differently.}
  \label{fig:robots}
\end{figure}

All levels of an iteration share one MuJoCo heightfield: the arena is a
grid of $16\times16$ tiles separated by flat aprons, each level is written
into its own tile and its environments spawn at the tile's centre. The
heightfield data is an array leaf of the MJX model, so writing 256 new
levels needs no recompilation, and all environments share one model.
Figure~\ref{fig:terrains} shows each feature at the end of its range and
Figure~\ref{fig:robots} the five robots on one level.

\FloatBarrier
\section{Methods and Hyperparameters}
\label{app:methods}

\paragraph{Training loop.} Brax's own training loop fixes each
environment's model for a whole run, so we drive brax's PPO networks, loss
and running observation normaliser from a custom loop. One iteration runs
256 lanes $\times$ 32 environments. All environments of a lane are reset
onto its level, with a fresh physics-randomisation draw where the robot
defines one, and stepped for one
episode in 20-step unrolls, with a PPO update after each unroll when the
iteration trains. An iteration is $256\cdot32\cdot300\approx2.46$M
environment steps (Spot: 6.4M).

\paragraph{Scores.} \emph{Positive value loss} (PVL): the mean positive
GAE advantage of a lane over the iteration~\citep{jiang2021plr}.
\emph{Learnability}: $p(1-p)$ with $p$ the fraction of the lane's
environments that never fell and reached the success
distance~\citep{rutherford2024noregrets}. \emph{Learning progress}
(after LP-ACRL~\citep{li2026lpacrl}): $R$ is the lane's mean over its 32
environments of the undiscounted return of each environment's first
episode on the level (up to 300 steps, Spot 780, steps after a fall and
reset still train PPO but do not count toward $R$). A visit's score is
$R$ minus the $R$ of the level's previous visit, signed, and 0 on a
level's first visit, including a mutated child's. \emph{PAIRED regret}:
the best antagonist return minus the mean protagonist return on the lane,
clipped at 0.

\paragraph{DR, DR-easy, DR-mixed.} Every iteration trains on 256 fresh
levels: uniform (DR), uniform shrunk to $d=0.1$ (DR-easy, the
same easy generator as ACCEL), or shrunk by a per-level
$d\sim\mathcal{U}(0,1)$ (DR-mixed).

\paragraph{Terrain-CL.} The game-inspired curriculum
of~\citet{rudin2022learning}, adapted to lanes: all 32 environments of a
lane share one level, so promotion is decided per lane. At the start of
the run each lane draws a uniform genome and noise seed (its column),
kept for the whole run, and a row $r\sim\mathcal{U}\{0,\dots,5\}$
(legged\_gym's initial-level cap). Every iteration trains on all lanes,
each at its column shrunk toward flat by $d=r/10$, the same shrink as the
random evaluation sets. After the iteration a lane moves to $r+1$ when at
least 50\% of its environments succeeded. Otherwise to $r-1$ (clipped
at 0) when its mean distance from spawn is below half its mean commanded
planar displacement, integrated from the commanded velocities and
heading rate over the episode (legged\_gym: distance below half the
commanded speed times the episode length). A lane promoted past row 10
moves to a uniformly random row.

\paragraph{\plr~\citep{jiang2021replay} and SFL.} A buffer of the 4000
highest-scoring levels. Once it holds 256 levels, an iteration replays
with probability 0.8: it samples 256 levels by rank prioritisation
(temperature 0.3) mixed with staleness (weight 0.3), trains on them and
refreshes their scores. Otherwise it scores 256 fresh uniform levels
without training and adds them to the buffer. \plr\ scores by PVL. Our
SFL scores by learnability in the same buffer. This is an approximation
of SFL~\citep{rutherford2024noregrets}, which periodically rebuilds a
top-$K$ buffer from a large random pool and trains on a mix of buffer and
random levels.

\paragraph{ACCEL~\citep{parkerholder2022accel} and variants.} \plr\ with
two changes: new levels come from the easy generator, and every replayed
level is mutated by perturbing 2 randomly chosen parameters by
$\mathcal{U}(-0.1,0.1)$ (clipped to $[0,1]$, seed kept). The next
non-training iteration scores the children of the most recent replay
batch. A non-training iteration with no queued children scores fresh
easy levels. ACCEL-grow adds a
``grow'' step that scales the mutated genome's distance from flat by
$\mathcal{U}(1,1.3)$. ACCEL-SFL is ACCEL-grow scored by learnability.
ACCEL-LP is ACCEL scored by learning progress. With signed scores,
levels whose return fell since their last visit rank last and are the
first evicted from a full buffer. LP-ACRL samples a fixed task set by a
softmax over learning progress. ACCEL-LP keeps ACCEL's rank and staleness
sampling over an evolving buffer.

\paragraph{PAIRED~\citep{dennis2020paired} and variants.} A learned
generator maps a 16-dimensional Gaussian latent through an MLP to a
sigmoid-squashed Gaussian over the 8 parameters. A random seed is
appended. Each iteration it proposes 256 levels, on which a protagonist
and an antagonist (two independent students, each with its own
normaliser) both train in separate rollouts. The generator is trained on
PAIRED regret by single-step PPO (entropy coefficient 0.1, learning rate
$10^{-4}$). UniBC and BiBC~\citep{mediratta2023stabilizing} add a
behaviour-cloning loss (weight 0.5, every iteration) that distils the
antagonist into the protagonist, or both ways. PAIRED+Evo replaces the
generator with ACCEL's buffer and edits, scored by PAIRED regret.

\paragraph{Budget accounting.} All methods get the same number of
environment steps. Non-training iterations of the buffer methods count
toward the budget, so they take fewer gradient steps than DR. PAIRED's
budget counts protagonist steps only, so PAIRED simulates twice as much
(Table~\ref{tab:compute}).

\begin{table}[!htbp]
  \centering\small
  \caption{Hyperparameters shared by all methods.}
  \label{tab:hparams}
  \begin{tabular}{ll}
    \toprule
    Lanes $\times$ environments per lane & $256\times32$ \\
    Steps per level (episode) & 300 (Spot 780) \\
    Environment steps & 300M (W01-TEK), 1B (others) \\
    PPO & brax, 20-step unroll, 4 epochs $\times$ 32 minibatches \\
    $\gamma$, $\lambda$, learning rate, entropy, clip, grad norm & 0.97, 0.95, $3\times10^{-4}$, 0.01, 0.2, 1.0 \\
    Networks & MLP 512-256-128, ELU, observation normalisation \\
    Buffer & 4000 levels, replay probability 0.8 \\
    Replay sampling & rank (temperature 0.3) + staleness (0.3) \\
    Easy generator & shrink to $d=0.1$ \\
    Terrain-CL & rows $d=r/10$, $r\in\{0..10\}$, $r_0\sim\mathcal{U}\{0..5\}$, up at $\ge$50\% success, \\
    & down at travel $<0.5\times$ commanded displacement \\
    Mutation & 2 parameters, $\pm0.1$, grow $\times\,\mathcal{U}(1,1.3)$ \\
    PAIRED generator & latent 16, lr $10^{-4}$, entropy 0.1 \\
    Behaviour cloning & weight 0.5, every iteration \\
    \bottomrule
  \end{tabular}
\end{table}

\FloatBarrier
\section{Evaluation Protocol}
\label{app:protocol}

\paragraph{Episodes.} Each final policy is evaluated deterministically
(tanh of the policy mean), without physics randomisation and without
early episode cuts. Commands are drawn as in training, with $v_x$
restricted to the evaluation range (App.~\ref{app:threshold}). Every level is rolled out in two independent
passes of 4 episodes each.

\paragraph{Level sets.} \emph{Random}: 2048 uniform genomes, shrunk to
$d\in\{0.25,0.5\}$. The genomes are the same at both.
\emph{Uniform}: a second, independent draw of 2048 uniform genomes, the
training distribution of DR. \emph{Flat}: 64 flat levels.
\emph{Held-out suite}: seven structured terrain families, each at strength
$k=1..4$ with 64 instances, radially symmetric like the training levels
(reference units, $\times s$): stairs up and down ($2k$\,cm steps every
0.3\,m), slopes up and down ($5k^\circ$), trenches (10\,cm deep, $5k$\,cm
wide, every 0.5\,m), waves ($\pm2k$\,cm, 0.6\,m wavelength) and random
blocks ($\pm2k$\,cm, 0.3\,m cells). We report mean success over the 28
cells.

\paragraph{Cross-fitted CVaR.} CVaR$_{25}$ is the mean success over the
worst 25\% of levels, after the
adversarial evaluation of~\citet{rutherford2024noregrets}. Levels are
ranked on one pass and scored on the other, in both directions, and
averaged. In a simulation with 2000 levels of true success 0.5 and 4
episodes each, same-pass CVaR$_{25}$ reads 0.19 and the cross-fitted
estimate 0.49.

\paragraph{Solvable subset.} Uniform success restricted to levels that
some policy passes: a level is solvable if at least one run of any method
on that robot succeeds in at least one episode of pass A. Every run is
scored on pass B over those levels and vice versa, and the two are
averaged. The subset depends on the set of evaluated runs, so the metric
is comparable only within one robot.

\paragraph{Capability envelopes.} For each of seven features (rough,
slope up, slope down, stairs up, stairs down, gaps, blocks), the feature
takes six evenly spaced magnitudes up to the end of its range with every
other feature flat, 64 seeded levels per magnitude. The envelope is the
largest magnitude still passed with 50\% success (linearly interpolated,
0 if the smallest magnitude already fails). An envelope at the end of the
range is censored.

\paragraph{Domination comparison.} For methods A and B, a level's solve
rate is the fraction of its 40 episodes (5 seeds $\times$ 2 passes
$\times$ 4) that succeed. The heatmap counts levels in a $10\times10$
grid of (A, B) solve rates. We report the fractions of levels where A's
rate is higher, lower or equal, the mean difference $A-B$ with a 95\%
bootstrap interval over levels, and the mean $|A-B|$ against a noise
reference: per level, both rates are drawn as $\mathrm{Binomial}(n,p)/n$
at $p=(A+B)/2$, and the reference is the exact expectation of their
absolute difference, averaged over levels. A mean $|A-B|$ at the
reference means the two methods are indistinguishable level by level.

\paragraph{Statistics.} Per-seed metrics are averaged over five seeds.
Tables give the mean $\pm$ sample standard deviation, and per-robot
method comparisons use two-sided Welch $t$-tests over the per-seed means,
uncorrected for multiple comparisons. Aggregates over robots
follow~\citet{agarwal2021precipice}: per robot, scores are min-max
normalised over the method means (so the bounds depend on the set of
methods). The IQM is the 25\%-trimmed mean of the 25 robot-seed scores. Intervals are
95\% percentile stratified bootstrap intervals with robots as strata
(10\,000 resamples). The probability of improvement $P(X>Y)$ is, per
robot, the Mann-Whitney statistic $U/(nm)$ of $X$'s against $Y$'s seeds
(ties count one half), averaged over robots, with a stratified bootstrap
interval (2000 resamples).

\FloatBarrier
\section{Additional Results}
\label{app:results}

\begin{table}[p]
  \centering\small
  \caption{Success rate (\%) per robot and method, mean $\pm$ std over 5
  seeds. Bold: top group per robot and column (best mean or not
  significantly below it, Welch $t$-test, $p\ge0.05$). CVaR$_{25}$ is over the
  $d=0.25$ levels, cross-fitted.}
  \label{tab:detail}
  \resulttable{detail}{Detail table (\texttt{tables/detail.tex}).}
\end{table}

\subsection{Full-difficulty (uniform) levels}
\label{app:uniform}

\begin{table}[!htbp]
  \centering\small
  \caption{Success rate (\%) on 2048 uniform levels: mean, cross-fitted
  CVaR$_{25}$ and the solvable subset (App.~\ref{app:protocol}), mean
  $\pm$ std over 5 seeds. Bold: top group per robot and column (Welch
  $t$-test, $p\ge0.05$).}
  \label{tab:uniform}
  \resulttable{uniform}{Uniform table (\texttt{tables/uniform.tex}).}
\end{table}

\begin{figure}[!htbp]
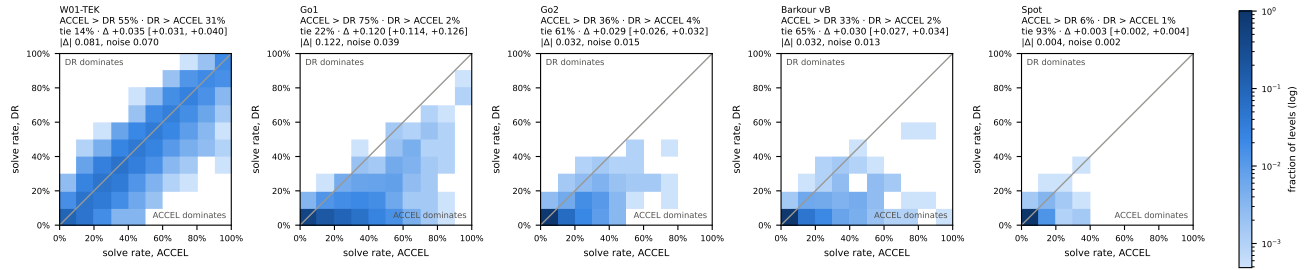

  \centering
  \resultfig{domination_param_uniform}{\linewidth}{Domination on uniform
  levels (\texttt{domination\_param\_uniform.pdf}).}
  \caption{ACCEL vs.\ DR, level by level, on the 2048 uniform levels. Read
  as Figure~\ref{fig:domination}.}
  \label{fig:domination-uniform}
\end{figure}

\FloatBarrier
\subsection{All method pairs}
\begin{figure}[!htbp]
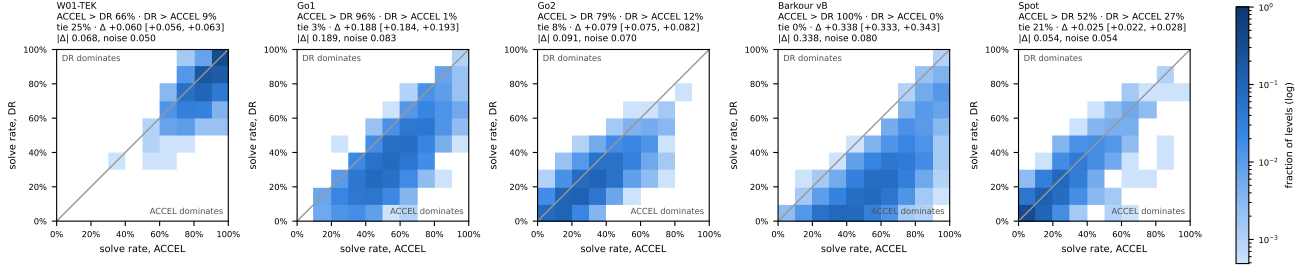

  \centering
  \resultfig{domination_accel_vs_dr}{\linewidth}{Domination heatmaps
  (\texttt{domination\_accel\_vs\_dr.pdf}).}
  \caption{ACCEL vs.\ DR, level by level, on the 2048 $d=0.25$ levels.
  Each level's solve rate is pooled over 5 seeds $\times$ 8 episodes. The
  colour is the fraction of levels (log scale) per cell. Mass below the
  diagonal: levels ACCEL solves more often. Panel titles give the
  fractions of levels ACCEL wins, loses and ties, the mean difference with
  its 95\% bootstrap interval, and the mean absolute difference against
  the binomial noise reference.}
  \label{fig:domination}
\end{figure}

\begin{figure}[!htbp]
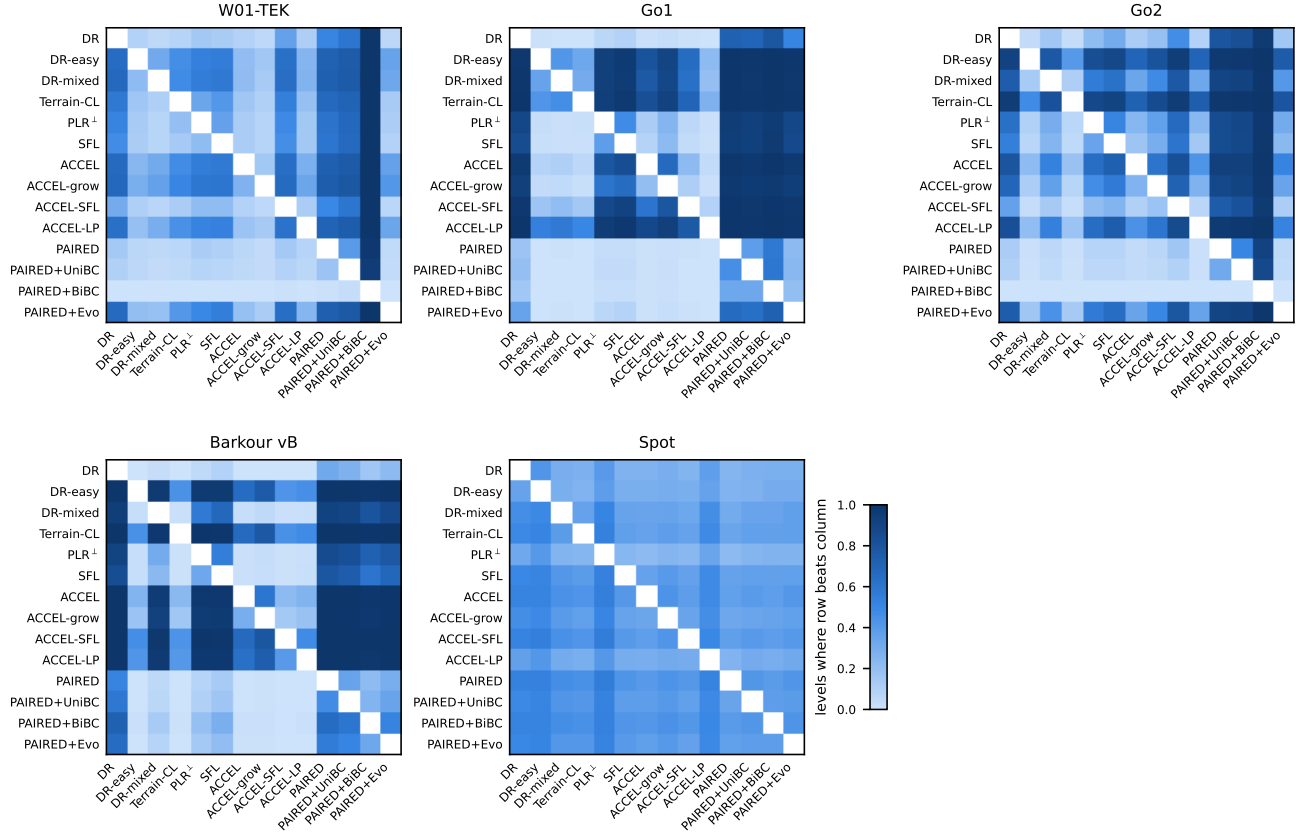

  \centering
  \resultfig{domination_matrix}{\linewidth}{Domination matrix
  (\texttt{domination\_matrix.pdf}).}
  \caption{Fraction of $d=0.25$ levels on which the row method's solve
  rate exceeds the column method's, per robot. Darker cells: the row method wins on more levels. The
  diagonal is blank, and $1-M_{ab}-M_{ba}$ is the fraction of ties.}
  \label{fig:matrix}
\end{figure}

The domination comparison (\S\ref{sec:protocol}) shows whether one
method is better on most levels or the two win on different levels.
Figure~\ref{fig:domination} gives it for ACCEL and DR on $d=0.25$ as an
example (Figure~\ref{fig:domination-uniform} on uniform levels), and
Figure~\ref{fig:matrix} for every pair.

\FloatBarrier
\subsection{Aggregates over robots}
\begin{figure}[!htbp]
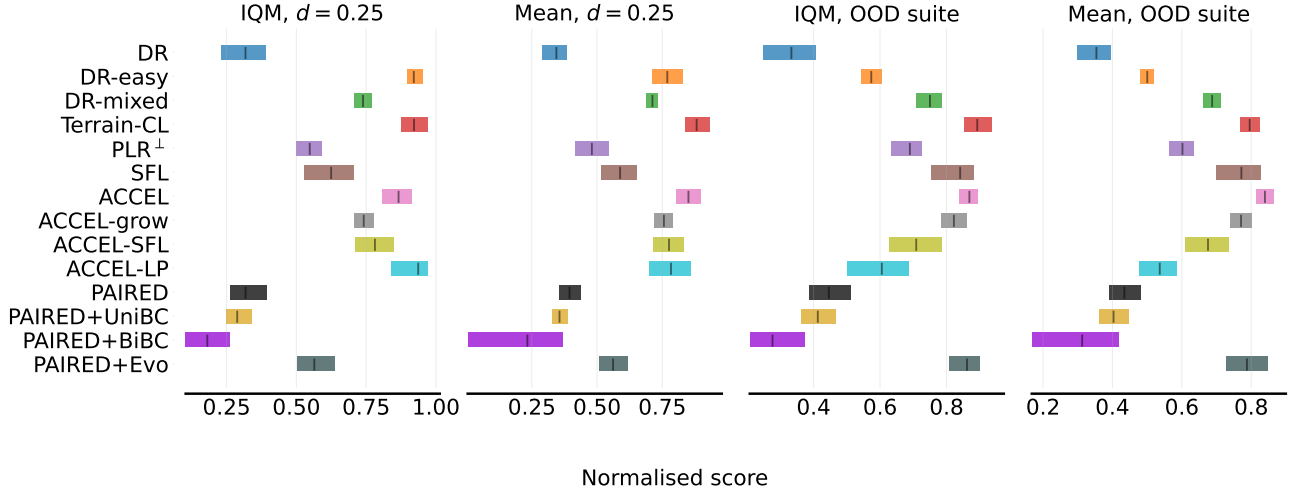

  \centering
  \resultfig{iqm}{\linewidth}{Aggregate IQM (\texttt{stats/iqm.pdf}).}
  \caption{Aggregate scores over five robots $\times$ five seeds: success
  min-max normalised per robot over the method means, interquartile mean
  and mean with 95\% stratified bootstrap intervals.}
  \label{fig:iqm}
\end{figure}

\begin{figure}[!htbp]
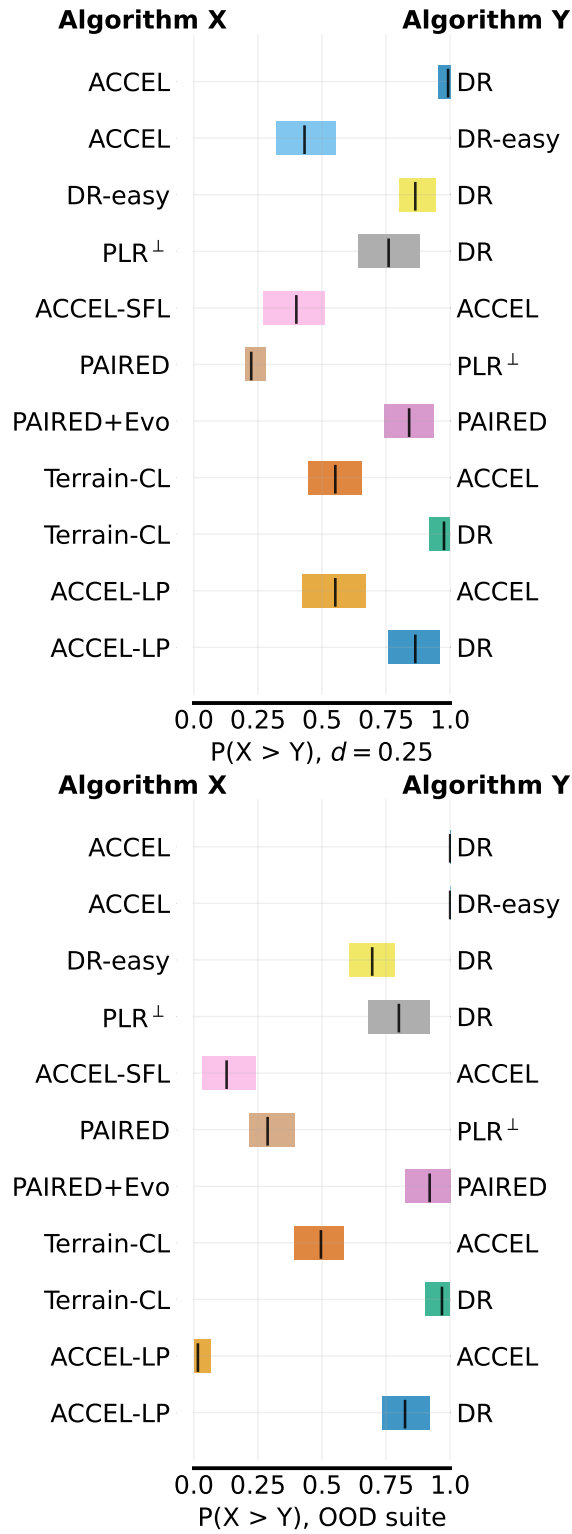

  \centering
  \resultfig{poi}{0.45\linewidth}{Probability of improvement
  (\texttt{stats/poi.pdf}).}
  \caption{Probability of improvement $P(X>Y)$ for the named method pairs,
  with 95\% stratified bootstrap intervals, on $d=0.25$ and the held-out
  suite.}
  \label{fig:poi}
\end{figure}

The main text compares methods within each robot. For completeness,
Figure~\ref{fig:iqm}, Table~\ref{tab:stats} and Figure~\ref{fig:poi}
pool the five robots
after min-max normalisation (\S\ref{sec:protocol}). By mean rank over
robots, ACCEL and Terrain-CL are first on $d=0.25$ (both 4.0) and on the
held-out suite (3.0 and 3.4), and DR is last on the suite (12.2). By
IQM (Table~\ref{tab:stats}), ACCEL-LP (0.94), DR-easy and Terrain-CL
(both 0.92) are first on $d=0.25$, and Terrain-CL (0.89), ACCEL (0.87)
and PAIRED+Evo (0.86) on the suite, with overlapping intervals. The
method rankings on $d=0.25$ agree across W01-TEK, Go1, Go2 and Barkour
(Spearman $\rho=0.53$-$0.86$, $p<0.05$ for all pairs but
W01-TEK-Barkour, $p=0.05$). Spot's ranking is unrelated to or inverted
against the others ($\rho=-0.61$ to $-0.28$).

\begin{table}[!htbp]
  \centering\small
  \caption{Aggregate IQM of normalised success over 5 robots $\times$ 5
  seeds with 95\% stratified bootstrap intervals (App.~\ref{app:protocol}).}
  \label{tab:stats}
  \resulttable{stats}{Stats table (\texttt{stats/stats.tex}).}
\end{table}

\FloatBarrier
\subsection{Capability envelopes}
\begin{table}[!htbp]
  \centering\small
  \caption{Capability envelopes: the largest feature magnitude passed at
  50\% success with every other feature flat, mean $\pm$ std over 5
  seeds. Bold: top group per robot and column (Welch $t$-test,
  $p\ge0.05$). In reference units (lengths are
  multiplied by $s$, slopes are not). An envelope at
  the end of the range (rough 10\,cm, slope $-20^\circ$ to $+18^\circ$,
  stairs $-8$ to $+5$\,cm, gaps 6\,cm, blocks 6\,cm) is censored.}
  \label{tab:envelopes}
  \resulttable{envelopes}{Envelope table (\texttt{tables/envelopes.tex}).}
\end{table}
On W01-TEK every method except PAIRED+BiBC reaches the end of the range
for rough ground, downhill slope and downward stairs. Stair-up height is the binding feature (DR
4.0, ACCEL 5.0 of 5\,cm). The open-chain robots' gap envelopes are at
most 2.7 of 6\,cm, and 0 for 42 of the 56 robot-method pairs.
Spot's uphill-slope envelope is at most
$5.2^\circ$.

\FloatBarrier
\subsection{Curriculum composition}
\begin{figure}[!htbp]
  \centering
  \resultfig{composition}{\linewidth}{Curriculum composition
  (\texttt{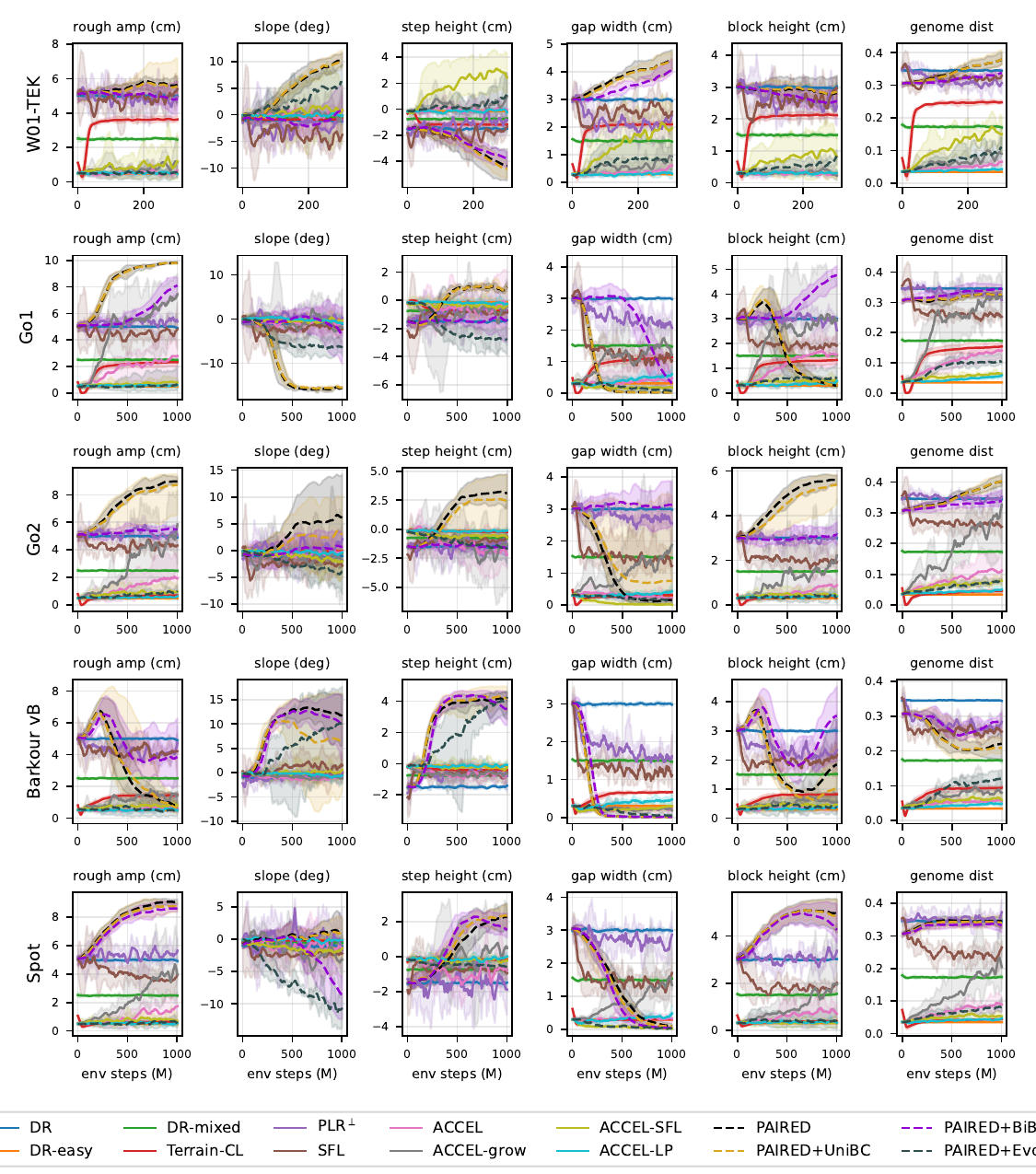}).}
  \caption{Mean physical value of each magnitude parameter among the
  levels the student trains on, and the mean genome distance from flat,
  vs.\ environment steps (reference units, running mean over 9 points). Mean
  over 5 seeds, shaded min-max.}
  \label{fig:composition}
\end{figure}

Figure~\ref{fig:composition} shows the terrain each curriculum trains on
over the run.

\FloatBarrier
\section{Compute}
\label{app:compute}
\begin{table}[!htbp]
  \centering\small
  \caption{Training compute per run, mean $\pm$ std over seeds. GPU-h: wall
  clock on one GPU. Sim steps count every simulated environment step
  (PAIRED's antagonist, the non-training iterations of the buffer
  methods). Student steps are those that fed a student PPO update.}
  \label{tab:compute}
  \resulttable{compute}{Compute table (\texttt{tables/compute.tex}).}
\end{table}
Every run used one NVIDIA H100. The 350 training runs took 306 GPU-hours
(W01-TEK 40, Go1 63, Go2 77, Barkour 66, Spot 60). A W01-TEK run (300M steps) takes 0.6
hours on average. A run of the other robots (1B steps, JAX backend) takes
0.9-1.1 hours. PAIRED and its variants take 1.8-1.9 times the wall clock of the
other methods.

\end{document}